\documentclass[10pt,twocolumn,letterpaper]{article}

\usepackage[pagenumbers]{cvpr} 

\usepackage{multirow}

\definecolor{cvprblue}{rgb}{0.21,0.49,0.74}
\usepackage[pagebackref,breaklinks,colorlinks,allcolors=cvprblue]{hyperref}

\def\confName{CVPR}
\def\confYear{2027}

\title{DocPristine: Content-Aware Document Appearance Enhancement under Real-World Mixed Degradations}

\author{
Zihan Huang$^{1}$ \quad Junle Liu$^{1}$ \quad Zhonghao Ye$^{1}$ \quad Yuyi Zhang$^{1}$ \quad Peirong Zhang$^{1}$\\
Hanlin Li$^{1}$ \quad Yongxin Shi$^{1}$ \quad Lianwen Jin$^{1}$\thanks{Corresponding author.}\\
{\normalsize $^{1}$South China University of Technology}\\
{\tt\small zihan\_huang416@foxmail.com \quad eelwjin@scut.edu.cn}
}

\begin{document}
\maketitle
\begin{abstract}
Camera-captured documents often exhibit mixed degradations, such as highlights, shadows, and blur, that distort backgrounds and obscure text details.
Document appearance enhancement (DAE) aims to mitigate these degradations and improve readability.
Existing document appearance enhancement methods rely either on input-derived visual cues that can become unreliable under severe degradation or on LLM-based degradation diagnosis that lacks spatial guidance.
Notably, we find that pretrained document visual encoders continue to attend to text regions in degraded documents, providing content-aware guidance.
Based on this insight, we propose \textbf{DocPristine}, which adaptively fuses multi-layer features from a frozen document visual encoder and injects them into a pixel restoration network.
To integrate these features into a pixel restoration network, DocPristine combines Adaptive Multi-Layer Feature Fusion (AMFF), Hierarchical Gated Conditioning (HGC), and Shifted-Window Cross-Attention (SWCA).
Specifically, SWCA draws on structural cues from neighboring locations across window boundaries to recover fine text details.
We also introduce \textbf{DocACE}, with a synthetic mixed-degradation training set (DocACE-Synthetic) and a real-world test set (DocACE-Real) containing manually restored ground-truth images and document parsing annotations.
Comprehensive experiments across three benchmarks demonstrate that DocPristine achieves state-of-the-art appearance enhancement quality and consistently improves downstream document parsing.
Code, models, and data will be publicly available.
\end{abstract}    

\begin{figure}[t]
    \centering
    \includegraphics[width=0.9\linewidth]{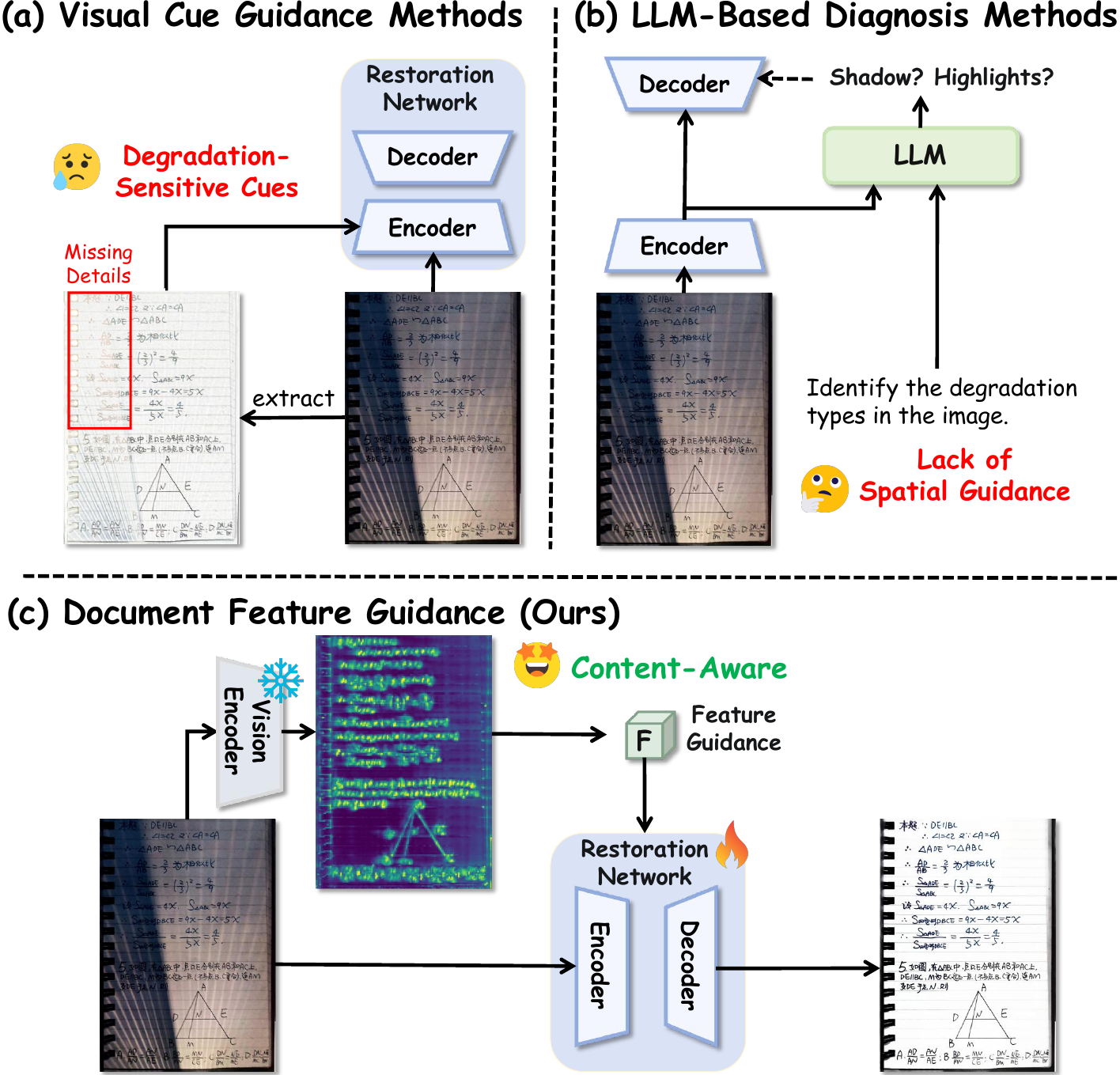}
    \caption{Comparison of Document Appearance Enhancement methods: (a) visual cue guidance methods, (b) LLM-based degradation diagnosis, and (c) DocPristine (ours).}
    \label{fig:motivation}
    \vspace{-5mm}
\end{figure}

\section{Introduction}
\label{sec:intro}
Photographed documents often exhibit multiple degradations, including shadows, highlights, color casts, and blur. These degradations can vary in severity across a page and overlap within the same region~\cite{zhang2023appearance,li2026mmdir}. 
Document Appearance Enhancement (DAE) aims to mitigate these degradations to improve readability and facilitate digital archiving~\cite{feng2021doctr,zhang2023document,li2023high,agam2007degraded}. 
A central challenge is to address these multiple degradations while preserving and recovering visual details such as character shapes, the spatial arrangement of symbols in formulas, and table lines. These details help downstream parsers recognize document elements and interpret their structure.


When multiple degradations coexist within a document image, existing DAE methods confront several challenges. 
\textbf{(1) Joint restoration under multiple degradations remains challenging.}
DocRes~\cite{zhang2024docres} and Uni-DocDiff~\cite{zhao2025uni} perform one restoration task at a time. Applying such methods to mixed degradations can require cascading multiple restoration operations. When degradations overlap, removing one can alter or obscure the visual cues needed to address another. As a result, the final output can depend on the order of operations, and errors introduced early can propagate through later stages.
\textbf{(2) Low-level visual cues can miss fine text details.}
DocRes~\cite{zhang2024docres} uses differences between the input image and its estimated background as a visual cue, and Uni-DocDiff~\cite{zhao2025uni} incorporates gradient and edge maps into its prior pool. These low-level visual cues are sensitive to the degradations they contain. Where highlights and blur overlap, text becomes harder to distinguish from the background, making these cues less informative for recovering fine text details.
\textbf{(3) LLM-based degradation diagnosis lacks spatial guidance.}
MMDIR~\cite{li2026mmdir} uses an LLM to identify different types of degradation, but offers limited spatial guidance on where text details need to be preserved or recovered.
This raises a natural question: \textit{Can incorporating finer-grained, content-aware guidance into a document appearance enhancement model help preserve and recover fine text details under mixed degradations?}

\begin{figure}[t]
    \centering
    \includegraphics[width=0.96\linewidth]{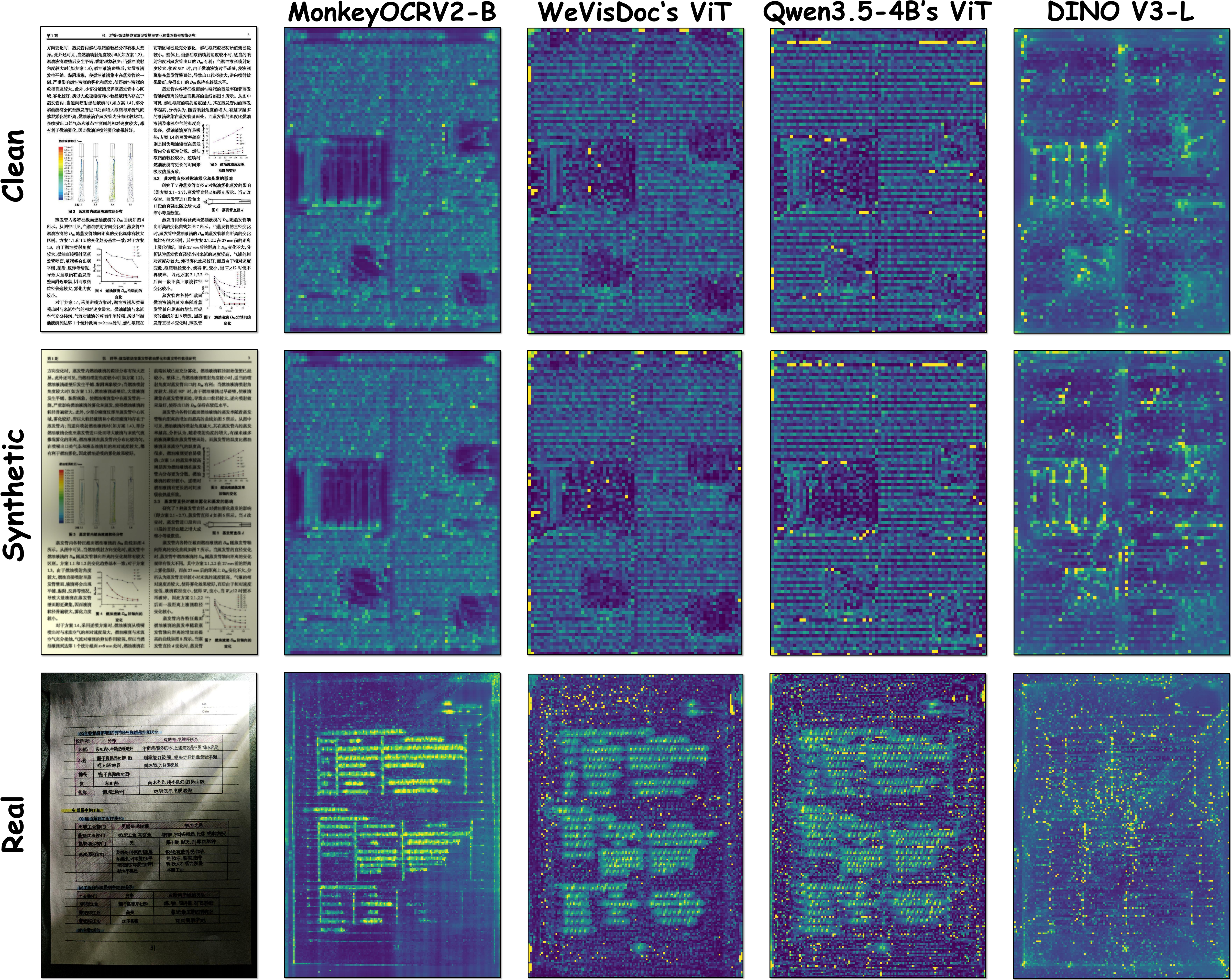}
    \vspace{-2mm}
    \caption{Visualization of attention maps from different visual encoders on clean and degraded document images.}
    \label{fig:attention_comparison}
    \vspace{-5mm}
\end{figure}

To identify a source of such guidance, we examine how pretrained visual encoders respond to degraded inputs. We find that existing document-adapted visual encoder (e.g., WeVisDoc-4B's ViT~\cite{wevisdoc}, MonkeyOCRv2-B~\cite{liu2026monkeyocrv2}) attends to text regions in clean documents and retains this focus in documents with synthetic and real-world degradations, as shown in Fig.~\ref{fig:attention_comparison}. 
Motivated by this observation, we propose \textbf{DocPristine}, a content-aware document appearance enhancement model for real-world mixed degradations. 
At its core, DocPristine uses features from a document-adapted visual encoder to guide background restoration and text detail recovery.
To fully exploit the hierarchical representations of a document-adapted visual encoder, we design \textbf{Adaptive Multi-Layer Feature Fusion (AMFF)} to adaptively combine shallow-layer appearance cues, such as color and intensity, with deeper-layer cues encoding text structure at each spatial location.
We further use \textbf{Hierarchical Gated Conditioning (HGC)} to adaptively inject the document guidance features into image features from the degraded input across locations and scales, helping restore background appearance and preserve text details.
Moreover, spatially adaptive feature injection alone may be insufficient when degradations severely obscure local text details. In such cases, neighboring regions with less severe degradations may provide additional structural cues. 
For example, when a highlight obscures part of a text line, nearby characters outside the highlight may still provide cues about character size and alignment.
To incorporate these cues into reconstruction, we introduce \textbf{Shifted-Window Cross-Attention (SWCA)}, which shifts the window partition so that restoration features can attend to document features at nearby locations across the original window boundaries. 
Together, these components help DocPristine better recover fine text details under real-world mixed degradations.

Due to the scarcity of DAE datasets covering the diverse combinations of degradations found in real-world captures, we introduce \textbf{DocACE}, which comprises a synthetic training set and a real-world test set.
Specifically, we construct \textbf{DocACE-Synthetic} using a carefully designed generation pipeline, which combines degradations of different types and severities to generate pixel-aligned pairs of degraded and clean images from 141,649 source document images. 
Furthermore, we introduce \textbf{DocACE-Real}, a benchmark of 120 photographed document pages under real-world conditions and exhibiting co-occurring degradations. To our knowledge, this is the first benchmark to provide expert-restored reference images and verified document parsing annotations for such photographs, enabling joint evaluation of restoration quality and downstream document parsing.

Extensive experiments on DocUNet, RealDAE, and DocACE-Real demonstrate that \textbf{DocPristine} achieves the highest PSNR and SSIM among the existing methods. On DocACE-Real, DocPristine improves PSNR by 3.42~dB and SSIM by 0.0603 over DocRes. With the same downstream document parser, DocPristine raises the overall parsing score from 78.61 on degraded inputs to 83.66 on restored images. 

Our main contributions are as follows.
\begin{itemize}
    \item We propose \textbf{DocPristine}, a content-aware document appearance enhancement method that leverages pretrained document visual features through three components (AMFF, HGC, and SWCA) to enhance document appearance under real-world mixed degradations while recovering fine text details.
    
    \item We introduce \textbf{DocACE}, comprising \textbf{DocACE-Synthetic} for training with pixel-aligned pairs covering diverse mixed degradations and \textbf{DocACE-Real}, the first benchmark to our knowledge to provide expert-restored references and verified document parsing annotations for real document photographs with mixed degradations.
    
    \item Extensive experiments demonstrate that \textbf{DocPristine} achieves state-of-the-art document appearance enhancement performance and significantly improves downstream document parsing performance.
\end{itemize}

\begin{figure*}[t]
    \centering
    \includegraphics[width=\textwidth]{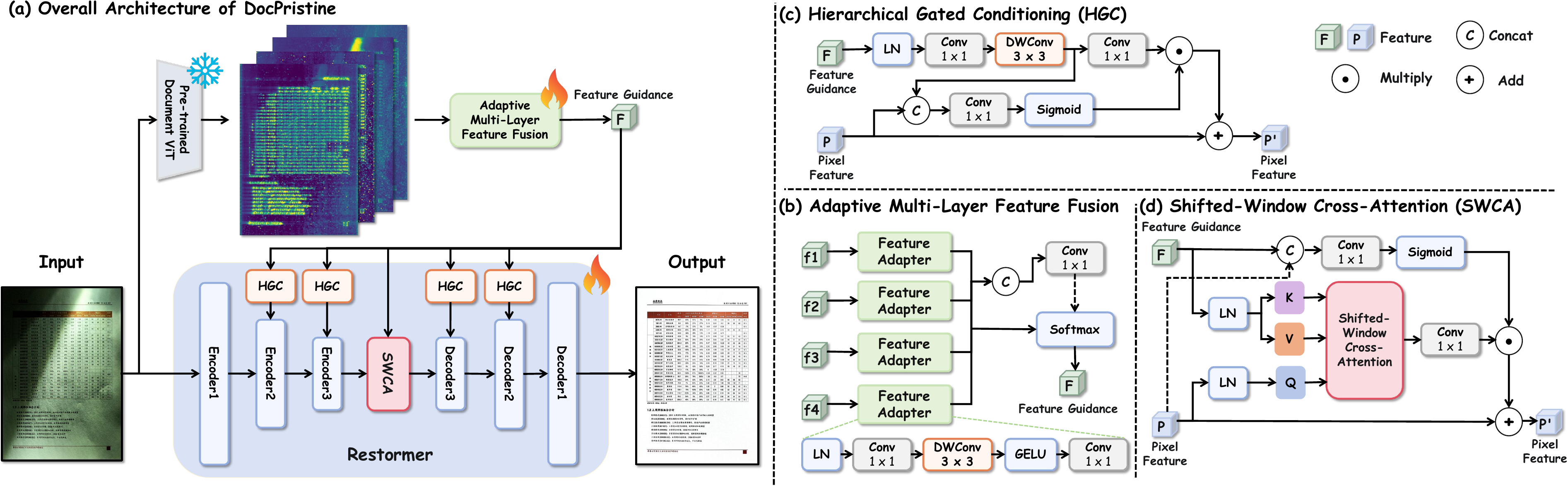}
    \caption{Overall architecture and key components of DocPristine, which integrates features from a document-adapted visual encoder into the restoration backbone through Adaptive Multi-Layer Feature Fusion (AMFF), Hierarchical Gated Conditioning (HGC), and Shifted-Window Cross-Attention (SWCA).}
    \label{fig:docpristine_architecture}
    \vspace{-3mm}
\end{figure*}

\section{Related Work}
\label{sec:related_work}

\subsection{Image Restoration under Mixed Degradations}
\label{sec:related_mixed_degradation}

Early image restoration methods typically address individual degradation types, with representative works on denoising, dehazing, deraining, and deblurring~\cite{zhang2017beyond,cai2016dehazenet,fu2017clearing,kupyn2018deblurgan}. To reduce reliance on task-specific models, some studies have explored all-in-one approaches that handle multiple degradation types within a unified framework. For instance, TransWeather~\cite{valanarasu2022transweather} introduces weather-type embeddings, AirNet~\cite{li2022all} learns degradation-aware representations, IDR~\cite{zhang2023ingredient} explores ingredient-oriented degradation modeling and PromptIR~\cite{potlapalli2023promptir} employs learnable prompts to condition the restoration process. However, these methods primarily focus on handling different degradation types within a unified framework, while the simultaneous occurrence of multiple degradations in a single image remains less explored. For example, AirNet~\cite{li2022all} considers multiple corruption types across a dataset, while each input image contains only one degradation. Recently, research has begun to explicitly address composite or complex degradations. OneRestore~\cite{guo2024onerestore} employs scene descriptors and cross-attention to guide restoration under composite weather conditions, while UniRes~\cite{zhou2025unires} combines specialized restoration models during diffusion sampling to handle arbitrary mixtures of known degradations. UnilDiff~\cite{cheng2026unildiff} further introduces degradation-aware feature fusion and detail-aware expert modules to handle diverse and compound degradations. Despite these advances, applying general image restoration methods to document images remains challenging, particularly in preserving fine text strokes, textual content, table lines, and layout.

\subsection{Document Appearance Enhancement}
\label{sec:related_document_enhancement}

Early work on document appearance enhancement primarily addressed illumination and color distortions introduced during capture.
DocProj~\cite{li2019document} applies illumination correction after geometric rectification, whereas DocIIW~\cite{das2020intrinsic} removes illumination effects through intrinsic decomposition.
GCDRNet~\cite{zhang2023appearance} further studies appearance enhancement for real photographed documents and introduces RealDAE with manually enhanced reference images.
However, document degradations extend beyond illumination and color distortions, and may involve co-occurring factors such as specular highlights, extensive shadows, paper-texture interference, and blur, with varying intensities and spatial distributions~\cite{li2026towardsrealworlddocument,li2026puredocbench,li2026mdpbenchbenchmarkmultilingualdocument,zhou2026real5omnidocbench}.

To address these challenges, recent methods have explored different guidance mechanisms. DocRes~\cite{zhang2024docres} constructs visual cues from task-specific priors extracted from degraded inputs, while Uni-DocDiff~\cite{zhao2025uni} combines learnable task prompts with spatial and frequency priors to guide restoration.
These low-level priors are themselves affected by input degradations, weakening their ability to guide text structure recovery.
For mixed-degradation document appearance enhancement, MMDIR~\cite{li2026mmdir} uses hidden representations from multimodal degradation identification responses to guide restoration and introduces the synthetic MixedDoc benchmark.
However, degradation identification focuses on characterizing image degradations and does not fully capture local text structures or their restoration requirements.
ENet-GP~\cite{burad2026enet} studies joint geometric and illumination correction using synthetic document data.
However, existing methods provide limited guidance for recovering local text structures while correcting complex appearance degradations. Therefore, simultaneously restoring background appearance and fine text details while preserving the original document geometry remains challenging.

\section{Method}
\label{sec:method}



\subsection{Motivation}
\label{sec:motivation}

Real-world mixed degradations can distort document backgrounds and obscure fine text details, reducing readability and affecting downstream document parsing. Document appearance enhancement aims to restore a clean, PDF-like appearance while preserving text content, layout, and geometry. Therefore, recovering fine-grained text structures while correcting overall appearance remains a key challenge.

Pretrained visual encoders offer a potential source of such guidance, as they can capture rich visual representations of document content. To investigate whether these representations retain attention to text regions under degraded conditions, we visualize and compare the final-layer self-attention maps of four pretrained ViT backbones:  MonkeyOCRv2-B~\cite{liu2026monkeyocrv2}, WeVisDoc-4B's ViT~\cite{wevisdoc}, Qwen3.5-4B's ViT~\cite{qwen3.5}, and DINOv3-L~\cite{simeoni2025dinov3}, across clean, synthetically degraded, and real-world documents with mixed degradations.
As shown in Fig.~\ref{fig:attention_comparison}, the encoders exhibit markedly different attention patterns. DINOv3-L pays limited attention to text regions, whereas document-adapted ViTs focus more strongly on these regions, particularly under mixed-degradation conditions.
These observations suggest that pretrained visual features, particularly those from document-adapted encoders, can provide spatially localized content cues for document restoration.

\subsection{DocPristine}
\label{sec:docpristine}


As shown in Fig.~\ref{fig:docpristine_architecture}, DocPristine uses Restormer~\cite{zamir2022restormer} as its image restoration backbone. It obtains content guidance from a frozen, pretrained document-adapted visual encoder, such as the visual encoder of WeVisDoc-4B~\cite{wevisdoc} or MonkeyOCRv2-B~\cite{liu2026monkeyocrv2}.
Different regions of a document may exhibit different types and degrees of degradation, so the cues needed for restoration vary across the page.
Adaptive Multi-Layer Feature Fusion (AMFF) combines complementary cues from different encoder depths at each spatial location. Hierarchical Gated Conditioning (HGC) regulates how much of the fused guidance is injected at each location and scale. Shifted-Window Cross-Attention (SWCA) draws on structural cues from neighboring locations, including those across the original window boundaries, to support fine text detail recovery. Together, these components adapt pretrained document features to the restoration needs of different regions.

\begin{figure*}[t]
    \centering
    \includegraphics[width=0.9\textwidth]{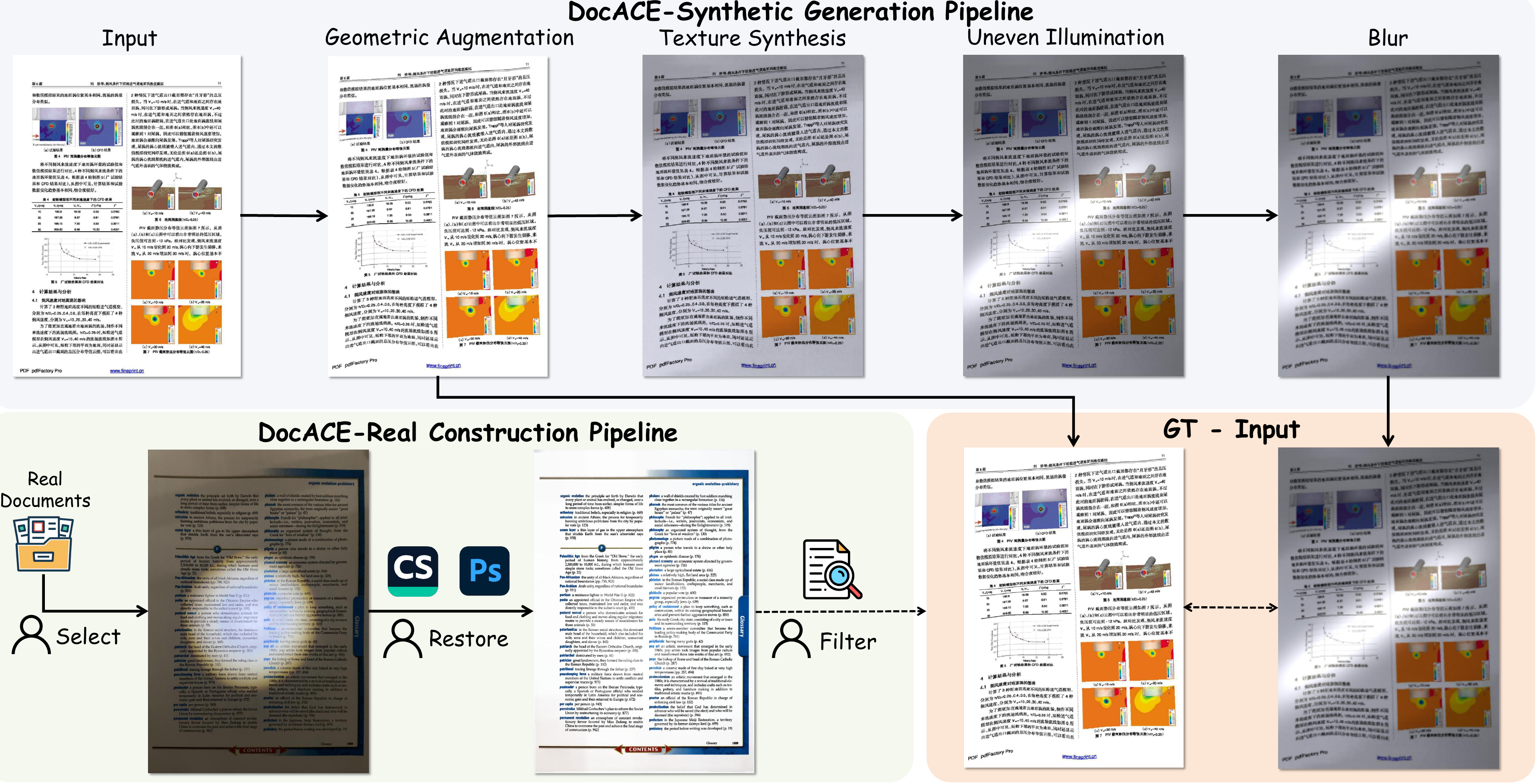}
    \vspace{-3mm}
    \caption{Overview of DocACE-Synthetic Generation Pipeline and DocACE-Real Construction Pipeline.}
    \vspace{-3mm}
    \label{fig:docace_pipeline}
\end{figure*}

\paragraph{Adaptive Multi-Layer Feature Fusion.}

Shallow features from the pretrained document-adapted visual encoder can retain color and intensity cues for background correction, while deeper features can provide structural cues for text detail recovery.
To address varying restoration needs across regions under real-world mixed degradations, we introduce AMFF, which learns fusion weights across encoder layers at each spatial location to adaptively integrate their complementary cues. 

Let $f_\ell\in\mathbb{R}^{768\times h\times w}$, $\ell\in\mathcal L=\{3,6,9,12\}$, denote the common-grid features.
Independent adapters $\mathcal A_\ell=W_{\ell,2}\circ\phi\circ D_\ell\circ W_{\ell,1}\circ\operatorname{LN}$ produce $Z_\ell=\mathcal A_\ell(f_\ell)$ with 128 channels.
With scores $s=W_\alpha[Z_3;Z_6;Z_9;Z_{12}]$ and spatial weights $\alpha=\operatorname{softmax}_\ell(s)$, fusion is

\begin{equation}
F=\sum_{\ell\in\mathcal L}\alpha_\ell\odot Z_\ell .
\label{eq:adaptive_layer_fusion}
\end{equation}

Throughout, $W$ denotes a $1\times1$ convolution, $D$ a $3\times3$ depthwise convolution, $\operatorname{LN}$ LayerNorm, $\phi$ GELU, $[;]$ channel concatenation, and $\odot$ elementwise multiplication with channel broadcasting.
The softmax normalizes layer weights at each location; $F$ guides HGC and SWCA.

\paragraph{Hierarchical Gated Conditioning.}
We introduce Hierarchical Gated Conditioning (HGC) to adapt the strength of document guidance across locations and scales, accommodating the different restoration needs of background and text regions.
HGC determines how much guidance to inject based on both the pixel features extracted from the degraded input and the fused document guidance features.

As shown in Fig.~\ref{fig:docpristine_architecture}(c), HGC is applied at the $1/2$- and $1/4$-resolution stages of both the encoder and decoder.
We bilinearly resize the document guidance features $F$ to match the spatial resolution of the pixel features $P$, then apply LayerNorm, a $1\times1$ convolution, and a $3\times3$ depthwise convolution to obtain the conditioning features:
\begin{equation}
C=D\!\left(W_c\!\left(\operatorname{LN}(\operatorname{Resize}(F))\right)\right).
\label{eq:hgc_conditioning}
\end{equation}
We concatenate $P$ and $C$ along channels and apply a $1\times1$ convolution followed by sigmoid to predict a spatial gate.
The gate weights the projected conditioning features for residual fusion:
\begin{equation}
P'=P+\sigma\!\left(W_g[P;C]\right)\odot W_o(C).
\label{eq:hgc}
\end{equation}
Through HGC, pixel features incorporate adaptive document guidance while retaining their original information, supporting text detail recovery and background correction.

\paragraph{Shifted-Window Cross-Attention.}
We apply Shifted-Window Cross-Attention (SWCA) at the bottleneck of the restoration network to incorporate structural cues from neighboring locations. Within each window, pixel features serve as queries, while the fused document features provide the keys and values. This allows each pixel feature to aggregate relevant document cues from nearby locations in the same window. Shifting the window partition brings neighboring locations on opposite sides of an original window boundary into the same attention window, allowing their structural cues to contribute to fine text detail recovery.

As shown in Fig.~\ref{fig:docpristine_architecture}(d), SWCA enables interactions between pixel features $P$ and fused document guidance features $F$ after the Restormer blocks at the bottleneck.
We first bilinearly resize $F$ to a lower-resolution key/value grid to obtain $F_k$.
LayerNorm and $1\times1$ convolutions map $P$ to queries $Q$ and $F_k$ to keys $K$ and values $V$.

Within each local window $w$, cross-attention aggregates document guidance features:

\begin{equation}
U_w=\operatorname{softmax}\!\left(\frac{Q_wK_w^\top}{\sqrt{d_h}}+B+M_w\right)V_w,
\label{eq:swca}
\end{equation}

where $d_h$ is the dimension of each attention head, $B$ is a learnable relative position bias, and $M_w$ is the shifted-window mask, which is zero for regular windows.
Regular and shifted windows enable interactions across window boundaries.
We reassemble the window outputs and restore their spatial positions to obtain features $U$.

We then upsample $F_k$ to the spatial resolution of $P$ to obtain $F_q$ for gating.
Concatenating $P$ and $F_q$, we apply a $1\times1$ gate convolution $W_g$ followed by the sigmoid function $\sigma$ to predict a spatial gate.
The gate weights the aggregated features projected by the $1\times1$ output convolution $W_o$ for residual fusion:
\begin{equation}
P'=P+\sigma\!\left(W_g[P;F_q]\right)\odot W_o(U).
\label{eq:swca_fusion}
\end{equation}
As in HGC, $P$ and $P'$ denote the input and output pixel features, while $W_g$ and $W_o$ denote the convolutions for gate prediction and output projection, respectively.

\paragraph{Loss Function.}
We train the model using a combination of L1 and SSIM losses:
\begin{equation}
\mathcal L=\|I_{\mathrm r}-I_{\mathrm{gt}}\|_1+\lambda\left[1-\operatorname{SSIM}(I_{\mathrm r},I_{\mathrm{gt}})\right],
\label{eq:training_objective}
\end{equation}
where $I_{\mathrm r}$ and $I_{\mathrm{gt}}$ denote the restored image and the clean target, respectively, and $\lambda=2$.

\begin{table}[t]
  \centering
  \caption{Comparison of document appearance enhancement benchmarks in terms of PDF-like restoration target, real-world captures, mixed degradations, and OCR evaluation.}
  \vspace{-2mm}
  \label{tab:benchmark_comparison}
  \scriptsize
  \setlength{\tabcolsep}{2pt}
  \newcommand{\benchmarkon}{\textcolor{ForestGreen}{$\checkmark$}}
  \newcommand{\benchmarkoff}{\textcolor{red}{$\times$}}
  \begin{tabular*}{\columnwidth}{@{\extracolsep{\fill}}lcccc@{}}
    \toprule
    \textbf{Benchmark} & \shortstack{\textbf{PDF-like}\\\textbf{Target}} & \shortstack{\textbf{Real}\\\textbf{Captures}} & \shortstack{\textbf{Mixed}\\\textbf{Degradations}} & \shortstack{\textbf{OCR}\\\textbf{Metrics}} \\
    \midrule
    DocUNet~\cite{ma2018docunet} & \benchmarkon & \benchmarkon & \benchmarkoff & \benchmarkoff \\
    RealDAE~\cite{zhang2023appearance} & \benchmarkon & \benchmarkon & \benchmarkoff & \benchmarkoff \\
    MixedDoc~\cite{li2026mmdir} & \benchmarkoff & \benchmarkoff & \benchmarkon & \benchmarkon \\
    \midrule
    \textbf{DocACE-Real (Ours)} & \benchmarkon & \benchmarkon & \benchmarkon & \benchmarkon \\
    \bottomrule
  \end{tabular*}
  \vspace{-4mm}
\end{table}

\begin{table*}[t]
  \centering
  \caption{Quantitative comparison of document appearance enhancement methods on DocUNet, RealDAE, and DocACE-Real. The best results are in \textbf{bold}, and second-best results are \underline{underlined}.}
  \vspace{-3mm}
  \label{tab:main_results}
  \small
  \setlength{\tabcolsep}{3pt}
  \begin{tabular*}{\textwidth}{@{\extracolsep{\fill}}ll*{10}{c}@{}}
    \toprule
    \multirow{2}{*}{\textbf{Method}} & \multirow{2}{*}{\textbf{Venue}} & \multicolumn{2}{c}{\textbf{DocUNet}~\cite{ma2018docunet}} & \multicolumn{2}{c}{\textbf{RealDAE}~\cite{zhang2023appearance}} & \multicolumn{6}{c}{\textbf{DocACE-Real}} \\
    \cmidrule(lr){3-4} \cmidrule(lr){5-6} \cmidrule(lr){7-12}
    & & PSNR$\uparrow$ & SSIM$\uparrow$ & PSNR$\uparrow$ & SSIM$\uparrow$ & PSNR$\uparrow$ & SSIM$\uparrow$ & TxE$\downarrow$ & FCM$\uparrow$ & TDS$\uparrow$ & $\mathrm{OCR}_{\mathrm{Overall}}\uparrow$ \\
    \midrule
    Input & -- & 13.09 & 0.7195 & 12.26 & 0.8264 & 7.05 & 0.4872 & 0.2329 & 78.21 & 80.92 & 78.61 \\
    GT & -- & -- & -- & -- & -- & -- & -- & 0.1148 & 85.20 & 86.95 & 86.89 \\
    \midrule
    DocProj~\cite{li2019document} & TOG'19 & 14.71 & 0.7098 & 19.35 & 0.8684 & 8.24 & 0.4725 & 0.4415 & 47.61 & 47.26 & 50.24 \\
    DocIIW~\cite{das2020intrinsic} & BMVC'20 & 16.42 & 0.7276 & 19.87 & 0.8633 & 17.68 & 0.8026 & 0.1509 & 78.35 & 80.73 & 81.33 \\
    DocTr~\cite{feng2021doctr} & MM'21 & 15.78 & 0.7067 & 18.62 & 0.7925 & 15.26 & 0.5488 & 0.1741 & 74.85 & 75.64 & 77.69 \\
    UDoc-GAN~\cite{wang2022udoc} & MM'22 & 14.29 & 0.6833 & 16.43 & 0.7558 & 12.18 & 0.5679 & 0.2798 & 49.51 & 65.40 & 62.31 \\
    GCDRNet~\cite{zhang2023appearance} & TAI'23 & 17.09 & 0.7658 & 24.42 & 0.9423 & 17.67 & 0.7754 & 0.1882 & 76.02 & 80.37 & 79.19 \\
    DocRes~\cite{zhang2024docres} & CVPR'24 & 17.60 & 0.7598 & 24.65 & 0.9219 & \underline{18.44} & \underline{0.8537} & \underline{0.1434} & \underline{79.78} & \underline{82.35} & \underline{82.60} \\
    Uni-DocDiff~\cite{zhao2025uni} & MM'25 & \underline{18.22} & \underline{0.7682} & \underline{24.97} & \underline{0.9485} & -- & -- & -- & -- & -- & -- \\
    ENet-GP~\cite{burad2026enet} & arXiv'26 & 15.70 & 0.7161 & 16.38 & 0.7621 & -- & -- & -- & -- & -- & -- \\
    \midrule
    \textbf{DocPristine (Ours)} & -- & \textbf{18.24} & \textbf{0.8032} & \textbf{25.01} & \textbf{0.9499} & \textbf{21.86} & \textbf{0.9140} & \textbf{0.1413} & \textbf{80.88} & \textbf{84.22} & \textbf{83.66} \\
    \bottomrule
  \end{tabular*}
  \vspace{-3mm}
\end{table*}

\begin{table}[t]
  \centering
  \caption{Comparison of visual encoders for appearance enhancement guidance. The best results are in \textbf{bold}, and the second-best results are \underline{underlined}.}
  \vspace{-2mm}
  \label{tab:visual_guidance}
  \scriptsize
  \setlength{\tabcolsep}{1pt}
  \resizebox{\columnwidth}{!}{%
  \begin{tabular}{@{}lc*{7}{c}@{}}
    \toprule
    \multirow{2}{*}{\shortstack{\textbf{Visual}\\\textbf{Encoder}}} & \multirow{2}{*}{\shortstack{\textbf{Params}\\\textbf{(B)}}} & \multicolumn{2}{c}{\textbf{DocUNet}~\cite{ma2018docunet}} & \multicolumn{2}{c}{\textbf{RealDAE}~\cite{zhang2023appearance}} & \multicolumn{3}{c}{\textbf{DocACE-Real}} \\
    \cmidrule(lr){3-4} \cmidrule(lr){5-6} \cmidrule(lr){7-9}
    & & PSNR$\uparrow$ & SSIM$\uparrow$ & PSNR$\uparrow$ & SSIM$\uparrow$ & PSNR$\uparrow$ & SSIM$\uparrow$ & $\mathrm{OCR}_{\mathrm{Overall}}\uparrow$ \\
    \midrule
    None & -- & 17.27 & 0.7725 & 23.22 & 0.9181 & 19.31 & 0.8482 & 81.59 \\
    \midrule
    DINOv3-L~\cite{simeoni2025dinov3} & 0.3 & 17.57 & 0.7734 & 23.12 & 0.9173 & 20.75 & 0.8851 & 82.13 \\
    Qwen3.5-4B ViT~\cite{qwen3.5} & 0.3 & 17.58 & 0.7690 & 23.94 & 0.9260 & \underline{21.26} & 0.8856 & 82.06 \\
    WeVisDoc-4B ViT~\cite{wevisdoc} & 0.3 & \underline{17.88} & \underline{0.7826} & \underline{24.64} & \underline{0.9418} & 21.22 & \underline{0.9045} & \underline{83.06} \\
    MonkeyOCRv2-B~\cite{liu2026monkeyocrv2} & 0.1 & \textbf{18.24} & \textbf{0.8032} & \textbf{25.01} & \textbf{0.9499} & \textbf{21.86} & \textbf{0.9140} & \textbf{83.66} \\
    \bottomrule
  \end{tabular}%
  }
\end{table}

\begin{table}[t]
  \centering
  \caption{Ablation of HGC and SWCA. Green checkmarks and red crosses indicate enabled and disabled components. $\mathrm{OCR}_{\mathrm{Overall}}$ is the downstream document parsing score on DocACE-Real. The best results are in \textbf{bold}, and the second-best results are \underline{underlined}.}
  \vspace{-2mm}
  \label{tab:component_ablation}
  \scriptsize
  \setlength{\tabcolsep}{1.5pt}
  \newcommand{\ablationon}{\textcolor{ForestGreen}{$\checkmark$}}
  \newcommand{\ablationoff}{\textcolor{red}{$\times$}}
  \begin{tabular*}{\columnwidth}{@{\extracolsep{\fill}}*{9}{c}@{}}
    \toprule
    \multirow{2}{*}{\textbf{HGC}} & \multirow{2}{*}{\textbf{SWCA}} & \multicolumn{2}{c}{\textbf{DocUNet}~\cite{ma2018docunet}} & \multicolumn{2}{c}{\textbf{RealDAE}~\cite{zhang2023appearance}} & \multicolumn{3}{c}{\textbf{DocACE-Real}} \\
    \cmidrule(lr){3-4} \cmidrule(lr){5-6} \cmidrule(lr){7-9}
    & & PSNR$\uparrow$ & SSIM$\uparrow$ & PSNR$\uparrow$ & SSIM$\uparrow$ & PSNR$\uparrow$ & SSIM$\uparrow$ & $\mathrm{OCR}_{\mathrm{Overall}}\uparrow$ \\
    \midrule
    \ablationoff & \ablationoff & 17.27 & 0.7725 & 23.22 & 0.9181 & 19.31 & 0.8482 & 81.59 \\
    \ablationon & \ablationoff & 17.39 & 0.7850 & 24.20 & 0.9215 & 20.63 & 0.8892 & 82.66 \\
    \ablationoff & \ablationon & \underline{17.92} & \underline{0.7994} & \underline{24.89} & \underline{0.9462} & \underline{21.35} & \underline{0.9033} & \underline{83.09} \\
    \midrule
    \ablationon & \ablationon & \textbf{18.24} & \textbf{0.8032} & \textbf{25.01} & \textbf{0.9499} & \textbf{21.86} & \textbf{0.9140} & \textbf{83.66} \\
    \bottomrule
  \end{tabular*}
  \vspace{-3mm}
\end{table}

\section{DocACE Dataset}
\label{sec:docace}

\subsection{DocACE-Synthetic}
Given the scarcity of large-scale training data covering diverse mixed degradations in document images, we construct DocACE-Synthetic.
Specifically, we collect 141,649 source document images from DocLayNet~\cite{pfitzmann2022doclaynet}, Doraemon~\cite{doraemon_pdf_layout_chinese}, M6Doc~\cite{cheng2023m}, CDLA~\cite{li2021cdla}, and publicly available papers from arXiv, and generate pixel-aligned pairs of degraded and clean document images by randomly combining appearance degradations of different types and strengths on the fly.

To capture the spatial co-occurrence and combined effects of degradations on document backgrounds and text details, we design the synthesis pipeline as follows.
We first diversify page geometry using deformation fields from Doc3D~\cite{das2019dewarpnet} and random rotations, taking the resulting images as clean ground-truth images.
Starting from these clean ground-truth images, we generate degraded inputs by overlaying paper shadows and textures from Doc3DShade~\cite{das2020intrinsic} and constructing spatially varying illumination fields to simulate broad cast shadows, specular highlights, and color casts.
We further apply Gaussian blur or downsampling to degrade text edges and fine details.
The resulting degraded inputs are pixelwise aligned with their corresponding clean ground-truth images.
Pseudocode and detailed hyperparameters are provided in the Supplementary Material.

\subsection{DocACE-Real}
To evaluate document appearance enhancement under real-world mixed degradations, we construct DocACE-Real, comprising 120 pairs of degraded camera-captured document images and expert-restored reference images.
Samples are collected from Wild-OmniDocBench~\cite{li2026towardsrealworlddocument}, PureDocBench~\cite{li2026puredocbench}, and Real5-OmniDocBench~\cite{zhou2026real5omnidocbench}. Document processing experts select challenging samples exhibiting mixed degradations.
The surrounding backgrounds are then cropped out to reduce environmental interference, yielding the degraded input images.

\begin{table}[t]
  \centering
  \caption{Ablation of feature layer selection. Zero, one, and four checkmarks indicate the baseline without feature injection, single-layer guidance, and adaptive fusion of all four layers, respectively.}
  \vspace{-2mm}
  \label{tab:feature_layer_ablation}
  \scriptsize
  \setlength{\tabcolsep}{1pt}
  \newcommand{\ablationon}{\textcolor{ForestGreen}{$\checkmark$}}
  \newcommand{\ablationoff}{\textcolor{red}{$\times$}}
  \resizebox{\columnwidth}{!}{%
  \begin{tabular}{@{}*{11}{c}@{}}
    \toprule
    \multirow{2}{*}{$f_3$} & \multirow{2}{*}{$f_6$} & \multirow{2}{*}{$f_9$} & \multirow{2}{*}{$f_{12}$} & \multicolumn{2}{c}{\textbf{DocUNet}~\cite{ma2018docunet}} & \multicolumn{2}{c}{\textbf{RealDAE}~\cite{zhang2023appearance}} & \multicolumn{3}{c}{\textbf{DocACE-Real}} \\
    \cmidrule(lr){5-6} \cmidrule(lr){7-8} \cmidrule(lr){9-11}
    & & & & PSNR$\uparrow$ & SSIM$\uparrow$ & PSNR$\uparrow$ & SSIM$\uparrow$ & PSNR$\uparrow$ & SSIM$\uparrow$ & $\mathrm{OCR}_{\mathrm{Overall}}\uparrow$ \\
    \midrule
    \ablationoff & \ablationoff & \ablationoff & \ablationoff & 17.27 & 0.7725 & 23.22 & 0.9181 & 19.31 & 0.8482 & 81.59 \\
    \midrule
    \ablationon & \ablationoff & \ablationoff & \ablationoff & \textbf{18.29} & 0.8017 & \textbf{25.20} & \textbf{0.9529} & 21.30 & 0.9007 & 82.27 \\
    \ablationoff & \ablationon & \ablationoff & \ablationoff & 18.16 & 0.8024 & 24.96 & \underline{0.9515} & 21.43 & 0.9084 & 82.59 \\
    \ablationoff & \ablationoff & \ablationon & \ablationoff & 18.18 & \textbf{0.8033} & 24.97 & 0.9496 & 21.74 & \underline{0.9101} & 83.18 \\
    \ablationoff & \ablationoff & \ablationoff & \ablationon & 18.17 & 0.8030 & 24.92 & 0.9476 & \underline{21.85} & 0.9094 & \underline{83.52} \\
    \midrule
    \ablationon & \ablationon & \ablationon & \ablationon & \underline{18.24} & \underline{0.8032} & \underline{25.01} & 0.9499 & \textbf{21.86} & \textbf{0.9140} & \textbf{83.66} \\
    \bottomrule
  \end{tabular}%
  }
  \vspace{-3mm}
\end{table}

For each selected sample, experts use CamScanner~\cite{camscanner} and Adobe Photoshop~\cite{adobephotoshop} to create clean reference images.
These references are prepared to resemble high-quality scans while preserving the original text content and layout.
To ensure their quality, each reference is independently reviewed by another expert who was not involved in its enhancement. Samples with visible residual degradations or damaged text content are excluded.

The source benchmarks also provide document parsing annotations, which we retain in DocACE-Real to support downstream evaluation. As shown in Table~\ref{tab:benchmark_comparison}, existing benchmarks differ in their restoration targets and evaluation capabilities. DocUNet~\cite{ma2018docunet} and RealDAE~\cite{zhang2023appearance} primarily focus on background illumination correction and overall color restoration, with less emphasis on accurate text recovery under complex mixed degradations. Neither provides downstream OCR evaluation. MixedDoc~\cite{li2026mmdir} focuses on removing synthetic shadows, seals, text watermarks, and blur, rather than restoring documents to a PDF-like appearance. In contrast, DocACE-Real targets camera-captured documents with mixed degradations and provides expert-restored references with clean backgrounds, vivid colors, and high text contrast, together with document parsing annotations. This enables joint evaluation of visual restoration quality and downstream document parsing performance.

\begin{figure*}[t]
  \centering
  \includegraphics[width=\textwidth]{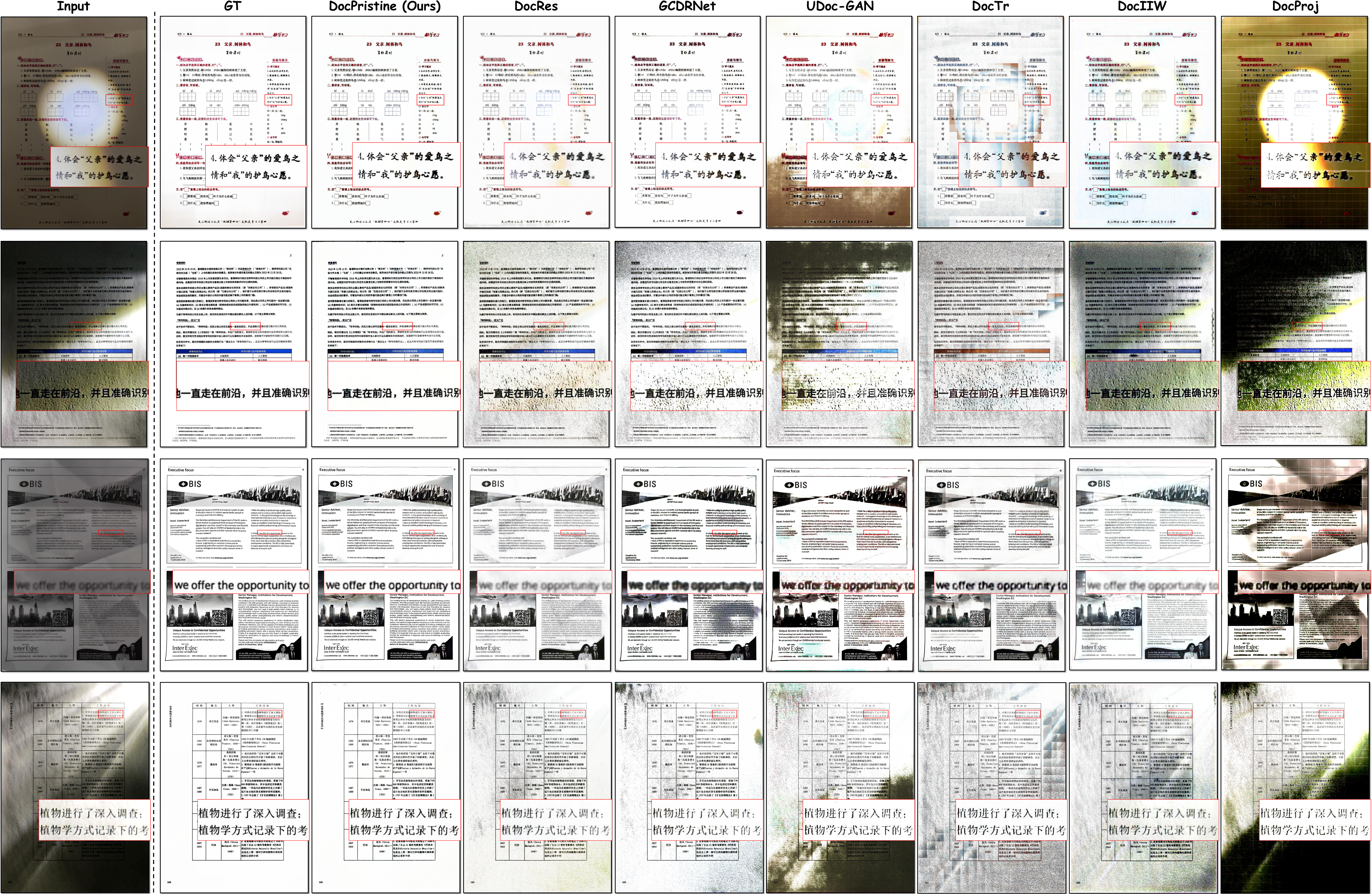}
  \caption{Comparison of results on DocACE-Real. Zoom in for a closer inspection of the appearance enhancement details.}
  \label{fig:qualitative_comparison}
  \vspace{-1mm}
\end{figure*}

\section{Experiments}
\label{sec:experiments}

\subsection{Experimental Settings}
\label{sec:experimental_settings}

\paragraph{Implementation Details.}
The model is trained for 10000 steps on four NVIDIA RTX A6000 GPUs using \(1024\times1024\) center crops and a batch size of 9 per GPU. The MonkeyOCRv2-B visual encoder remains frozen, while only the restoration network and the feature fusion and guidance modules are optimized using AdamW. Both the initial learning rate and weight decay are set to \(10^{-4}\). Following 200 steps of linear warmup, the learning rate decays to \(10^{-5}\) according to a cosine schedule. Training uses BF16 mixed precision, and testing employs full-image inference. All experiments share these training and inference settings.
We train the model on 141,649 pairs of degraded and clean document images from DocACE-Synthetic.


\paragraph{Evaluation Metrics.}
We adopt peak signal-to-noise ratio (PSNR) and structural similarity (SSIM)~\cite{wang2004image} as metrics.
Both metrics are computed for each image and averaged across images.
We further evaluate downstream performance on DocACE-Real using PaddleOCR-VL-1.6~\cite{zhang2026paddleocrvl16expandingfrontierdocument}.
Following the evaluation protocol of OmniDocBench~\cite{ouyang2024omnidocbenchbenchmarkingdiversepdf}, we report TextEdit (TxE), FormulaCDM (FCM), and TableTEDS (TDS).
TxE ranges from $0$ to $1$, while FCM and TDS are both reported on a $0$--$100$ scale.
We convert TxE to text similarity, $100(1-\mathrm{TxE})$, and take its equally weighted average with FCM and TDS to obtain the overall document parsing score, $\mathrm{OCR}_{\mathrm{Overall}}$:
\begin{equation}
\mathrm{OCR}_{\mathrm{Overall}}=\frac{1}{3}\left[100(1-\mathrm{TxE})+\mathrm{FCM}+\mathrm{TDS}\right].
\label{eq:parsing_overall}
\end{equation}
$\mathrm{OCR}_{\mathrm{Overall}}$ ranges from $0$ to $100$, with higher values indicating better performance.

\subsection{Main Results}
\label{sec:main_results}

We compare DocPristine with DocProj~\cite{li2019document}, DocIIW~\cite{das2020intrinsic}, DocTr~\cite{feng2021doctr}, UDoc-GAN~\cite{wang2022udoc}, GCDRNet~\cite{zhang2023appearance}, DocRes~\cite{zhang2024docres}, Uni-DocDiff~\cite{zhao2025uni}, and ENet-GP~\cite{burad2026enet} on 130 images from DocUNet~\cite{ma2018docunet} after alignment using DocAligner~\cite{zhang2026docaligner}, 150 images from the RealDAE~\cite{zhang2023appearance} validation set, and 120 images from DocACE-Real.
Uni-DocDiff~\cite{zhao2025uni}, ENet-GP~\cite{burad2026enet}, and MMDIR~\cite{li2026mmdir} are not evaluated on DocACE-Real because their models and code are not publicly available. 
On DocACE-Real, we evaluate competing methods using official implementations and publicly released models, following their original inference pipelines.
As shown in Table~\ref{tab:main_results}, DocPristine achieves the best reported PSNR and SSIM on all three benchmarks.
On DocUNet, it improves SSIM by 0.0350 over the second-best method, Uni-DocDiff.
On DocACE-Real, DocPristine achieves a PSNR of 21.86~dB and an SSIM of 0.9140, outperforming the second-best method, DocRes, by 3.42~dB and 0.0603, respectively.
These results demonstrate effective document restoration under real-world mixed degradations.

For downstream document parsing on DocACE-Real, DocPristine also achieves the best results on all four metrics.
It raises $\mathrm{OCR}_{\mathrm{Overall}}$ from 78.61 for the degraded inputs to 83.66, outperforming the second-best method, DocRes, by 1.06 points.
Compared with DocRes, DocPristine reduces text edit distance from 0.1434 to 0.1413 and improves formula recognition and table parsing scores by 1.10 and 1.87 points, respectively.
These results show that the gains in appearance enhancement quality also translate into improved downstream document parsing.
Fig.~\ref{fig:qualitative_comparison} presents qualitative comparisons on DocACE-Real.
The supplementary material compares DocPristine with DocRes~\cite{zhang2024docres}, GCDRNet~\cite{zhang2023appearance}, and a plain Restormer~\cite{zamir2022restormer} baseline, all trained on DocACE-Synthetic.

\subsection{Visual Guidance Analysis}
\label{sec:visual_guidance_analysis}

We compare four pretrained visual encoders as sources of feature guidance for DocPristine, including two general-purpose ViTs (DINOv3-L and Qwen3.5-4B) and two document-oriented ViTs (MonkeyOCRv2-B and WeVisDoc-4B).
For each encoder, we extract features from four layers evenly spaced across its depth and use the same multi-layer feature fusion and injection modules to guide the restoration network.
The attention analysis in Fig.~\ref{fig:attention_comparison} shows that these pre-trained document ViTs adapted to document tasks focus more strongly on text regions in the illustrated clean and degraded examples.

As shown in Table~\ref{tab:visual_guidance}, using either MonkeyOCRv2-B or WeVisDoc-4B improves PSNR and SSIM over the variant without feature injection on three benchmarks and yields a higher $\mathrm{OCR}_{\mathrm{Overall}}$ on DocACE-Real.
Both document ViTs outperform DINOv3-L and Qwen3.5-4B in SSIM and document parsing scores.
WeVisDoc-4B achieves an SSIM of 0.9045 and an $\mathrm{OCR}_{\mathrm{Overall}}$ of 83.06 on DocACE-Real, compared with 0.8482 and 81.59 without feature injection.
These results are consistent with the attention analysis and indicate that visual encoders adapted to document tasks provide effective content guidance for appearance enhancement.
MonkeyOCRv2-B achieves the best results across all appearance enhancement and parsing metrics, and we therefore select it as the visual guidance encoder for DocPristine.

\subsection{Ablation Study}
\label{sec:ablation_study}

We use the restoration network without feature injection as the baseline and enable HGC, SWCA, or both.
As shown in Table~\ref{tab:component_ablation}, either module improves appearance enhancement quality on all three benchmarks and the overall document parsing score on DocACE-Real.
Combining the two modules further improves performance, increasing PSNR, SSIM, and $\mathrm{OCR}_{\mathrm{Overall}}$ on DocACE-Real by 0.51~dB, 0.0107, and 0.57 points, respectively, over the variant using only SWCA.
These results demonstrate the complementary roles of the two guidance mechanisms.

With HGC and SWCA enabled, we compare guidance from encoder layers 3, 6, 9, and 12 individually with adaptive fusion of all four layers.
As shown in Table~\ref{tab:feature_layer_ablation}, guidance from each selected layer outperforms the baseline without feature injection, although the best single layer varies across benchmarks.
Layer 3 achieves the highest PSNR on DocUNet~\cite{ma2018docunet} and RealDAE~\cite{zhang2023appearance}, reaching 18.29~dB and 25.20~dB, respectively.
We attribute this difference to the restoration priorities of these benchmarks.
The ground-truth images in DocUNet and RealDAE primarily reflect corrections to background illumination and overall color, with less emphasis on fine text detail recovery.
Shallow features retain richer color and local appearance information, helping correct uneven background illumination and color casts introduced during capture, which translates into better pixel metrics.

DocACE-Real evaluates both appearance enhancement and content recovery accuracy under real-world mixed degradations.
Illumination and color distortions overlap with specular highlights, interference from paper texture, and blur, affecting both background appearance and text recognition and placing greater demands on feature guidance.
As the selected layer deepens from layer 3 to layer 12, PSNR increases from 21.30~dB to 21.85~dB, while the overall parsing score rises from 82.27 to 83.52.
These results indicate that deeper document features provide more effective guidance for accurate content recovery under real-world mixed degradations.
Adaptive fusion of all four layers achieves the best PSNR, SSIM, and overall parsing score, highlighting the complementary roles of shallow appearance information and deeper content representations in improving both visual quality and content readability.

\section{Conclusion}
\label{sec:conclusion}

DocPristine integrates features from a frozen pretrained document ViT through AMFF, HGC, and SWCA to restore text and backgrounds under real-world mixed degradations.
DocACE combines synthetic training pairs with a real-world benchmark containing expert-restored references and verified parsing annotations.
Experiments on DocUNet, RealDAE, and DocACE-Real demonstrate state-of-the-art appearance enhancement, with improved downstream document parsing on DocACE-Real.

\section*{LLM Usage Statement}
We used OpenAI's ChatGPT to refine author-written text, related work, content on dataset preparation, and the presentation of research ideas and conclusions.
The tools also provided reviewer-style feedback on the manuscript and generated implementation code that the authors reviewed and approved.
The authors take full responsibility for all submitted content.

{
    \small
    \bibliographystyle{ieeenat_fullname}
    \bibliography{main}
}

\end{document}